\pdfoutput=1
\documentclass{OAGM}
\OAGMarXiv{1404.3538}

\usepackage{graphics}
\usepackage{graphicx}

\usepackage{multirow}

\usepackage{tabularx}

\usepackage{amsmath,amssymb}   
\usepackage{subfigure}
\usepackage{sidecap}
\usepackage{authblk}

\title{Geometric Abstraction from Noisy Image-Based 3D Reconstructions}

\author[1,2]{Thomas Holzmann}
\author[1]{Christof Hoppe}
\author[2]{Stefan Kluckner}
\author[1]{Horst Bischof}

\affil[1]{Institute for Computer Graphics and Vision\\

 Graz University of Technology, Austria}
\affil[2]{Siemens AG Austria, CT RTC ICV }

\begin{document}
\maketitle

\vspace{-0.8cm}
\begin{abstract}
Creating geometric abstracted models from image-based scene reconstructions
is difficult due to noise and irregularities in the reconstructed model. 
In this paper, we present a geometric modeling method for noisy reconstructions 
dominated by planar horizontal and orthogonal vertical structures. We partition 
the scene into horizontal slices and create an inside/outside labeling 
represented by a floor plan for each slice by solving an energy minimization problem. 
Consecutively, we create an irregular discretization of the volume according
to the individual floor plans and again label each cell as inside/outside by 
minimizing an energy function. By adjusting the 
smoothness parameter, we introduce different levels of detail. 
In our experiments, we show results with varying regularization 
levels using synthetically generated and real-world data. 
  
\end{abstract}

\vspace{-0.6cm}
\section{Introduction}
\vspace{-0.4cm}
Thanks to recent advances in image-based 3D reconstruction techniques 
like Structure-from-Motion~\cite{snavely06}\,(SfM) and patch-based multi view stereo~\cite{furukawaPMVS}\,(PMVS) and
the availability of  depths sensors like Time-of-Flight cameras or with structured light technology, we can easily
create large-scale, high-resolution 3D reconstructions containing millions of
points. 

However, transmitting, visualizing, processing and analyzing the acquired data
is far from practical use within applications.
For example, to transmit a point cloud of millions of points over the Internet is
not possible in reasonable time and therefore abstracted models are desired. Further, to extract 
semantic information out of a 3D reconstruction, it has to
be separated into semantic meaningful parts first. For example, by separating
a building model into floors, a floor plan can be extracted for each level and
can be processed individually.

There exist several approaches for representing a 3D model in a simplified
way. In Wu et al. \cite{wu12}, a parametric method for reconstructing
architectural scenes from sparse point clouds is proposed. 
Profile curves are swept over a network of transport curves in order to
generate swept surfaces. 

Xiao and Furukawa~\cite{xiao12} introduce a method to automatically construct
and visualize 3D models for large indoor scenes using 3D laser scanner data as input.
Their approach partitions the 3D model into horizontal slices, which are 
bounded by dominant horizontal structures. For each slice, walls are detected
by projecting all laser points to the 2D ground plane and then using a Hough
transform to detect lines. Each floor plan is approximated by a 2D Constructive 
Solid Geometry (CSG) representation. To obtain a full 3D model, the 2D CSG
representation is lifted into 3D and again optimized to obtain consistent 3D models.
If the condition that the scene purely consists of vertical structures is violated or
the 3D data is perturbed by significant amount of noise, the wall detection
using Hough lines fails and the overall result degrades. Therefore this approach
is not suited for SfM results that contain significant amount of noise and
outliers. 

A similar approach for multi-level indoor scenes (i.e., whole buildings) which
is not limited to orthogonal or parallel structures is proposed by Oesau et
al.~\cite{oesau13}. First, permanent structures (wall, floor, ceiling) are
detected using horizontal slicing and wall directions are computed using Hough
transform. For every slice, a triangular decomposition is created and extruded
to 3D to form irregularly shaped volumetric cells. Finally, the volumetric cells are labeled as empty or occupied
space by optimizing an energy function that is modeled by a conditional random field (CRF). 
In comparison to~\cite{xiao12}, this approach improves the
handling of rounded vertical structures. Though, it also assumes laser data
with little noise as input data.

In this paper, we focus on extracting meaningful geometric structures of
man-made environments which are often dominated by planar surfaces. In
contrast to the work of~\cite{xiao12} and \cite{oesau13} that rely on clean
input data obtained by a laser scanner, we focus on surface meshes as they are
created by standard image-based reconstruction pipelines like
PMVS combined with meshing techniques like Poisson triangulation~\cite{kazhdan06} . Compared to laser scan data, the mesh has a
much higher level of noise and often contains artifacts which complicates the extraction of a simplified geometry.

Following the idea of~\cite{xiao12}, we first identify horizontal
slices that are limited by dominant horizontal planar structures. For each
slice, we generate a 2D floor plan by solving an inside/outside labeling
problem in a global optimal manner formulating an energy minimization problem. In order to integrate all
floor plans into a consistent 3D model, we create an irregular discretization
of the volume according to the individual floor plans. The obtained volume
elements are again labeled as inside/outside by minimizing an energy function. For an
individual building, the whole procedure results in a set of floor plans and an
adjustably regularized geometric abstraction of the input 3D point cloud. In
our experimental evaluation, we show that our approach massively
simplifies the input mesh while the results are geometrically consistent with
the input data. We demonstrate that our approach is computationally efficient
and delivers accurate results where existing methods may fail.

\vspace{-0.4cm}
\section{Our Approach}
\label{sec:approach}
\begin{figure}
\centering
\includegraphics[width=0.7\linewidth]{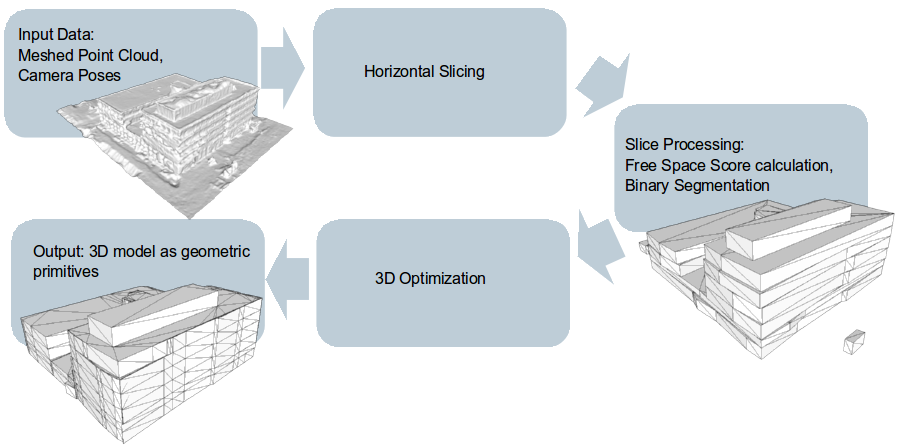}
\caption[Overview of our processing workflow.]{\textit{Overview of our processing workflow.} 
}
\label{fig:workflow}
\end{figure}

\vspace{-0.3cm}
Given a meshed point cloud and the corresponding camera poses $C$ as
input, we want to extract relevant geometric structures of the scene. As
precondition, we assume that it is dominated by planar horizontal and orthogonal vertical surfaces.
We apply preprocessing steps to transform 
the model to a new coordinate system aligned with a reference plane and 
perform a predefined, constant upsampling of the meshed point cloud.
At dominant horizontal structures, we partition the model into
horizontal slices. A slice represents a part of the model which includes mainly
vertical structures and is bounded by horizontal structures. For each slice, we
create a 2D floor plan using the visibility information. The individual
floor plans are used to create an irregular discretization of the
volume into cells. Finally, we obtain a regularized 3D model by labeling each
cell as inside/outside using a CRF.
Fig. \ref{fig:workflow} illustrates the different modules of our processing workflow.

\vspace*{-0.4cm}
\subsection{Slicing and Contour Extraction}
\label{sec:slicing}
\vspace{-0.2cm}
In this section, we describe the processing of the point cloud in the slice
domain. 
Initially, the model gets
partitioned into horizontal slices. A slice includes similar vertical
structures, e.g. vertical walls, and is bounded by horizontal structures. For
each slice we calculate a 2D binary segmentation resulting in a floor 
plan. The
segmentation is performed by using the visibility information that is associated
with the input mesh. Finally, we approximate the border of the
segmentation with line segments. 

\textbf{Slice Extraction.}
In our workflow, slices are defined as parts of the model enclosed by dominant
horizontal structures. We find these structures by projecting points with
normals similar to the ground plane normal, i.e. similar to the z-axis, onto
the z-axis. On this one-dimensional data, we perform mode estimation using mean shift~\cite{comaniciu02}. We use the centers of
the modes as boundaries of the slices. 

The mean shift bandwidth is calculated as a fraction of the model height. This
leads to independence of the model size and just the denominator $d$ has to be
defined: $bandwidth = \frac{height}{d}.$ 
By adjusting the bandwidth parameter of mean shift, it is possible to generate
slice boundaries at all minor horizontal structures or just at the most
important dominant structures. Thus, the parameter $d$ defines the potential
level of detail in the vertical direction of our modeling approach.

\textbf{Binary Segmentation.}
In order to identify the dominant vertical structures of each slice, we perform
a binary inside/outside segmentation in the discretized 2D slice
plane. The discontinuity between differently labeled regions is then the resulting 2D
floor plan. Visibility information is used to define a probability for
every pixel for being inside or outside of the
object. We call this probability free space score. Finally, we define an energy
function to obtain an optimal labeling.

To calculate the free space score for each pixel in 2D, we first create a voxel
grid spanned over the whole scene and assign each voxel a
score for being free and occupied space. This score is defined by 
the number of cameras a certain voxel is visible in.
Therefore, we cast rays from each voxel $v_{xyz}$ to all camera centers $C$. If
a ray from $v_{xyz}$ to a camera $c \in C$ does not intersect the input
mesh, $v_{xyz}$ is visible in $c$. The score that $v_{xyz}$ is in free space is defined as
\vspace{-0.2cm}
\begin{equation}
p(v_{xyz} = outside | visibility) = \frac{\{\#\ cameras\ v_{xyz}\ is\ visible\ in \}}{\{max\ \#\
visible\ cameras\}},
\end{equation}
\vspace{-0.5cm}

where $\{max\ \#\ visible\ cameras\}$ is the maximum number of visible 
cameras for a voxel $v_{xyz}$.
\\
For all the voxels $v_{xyz} \in V$ which have not been visible in any camera view,
we define the score that $v_{xyz}$ is in occupied space  by calculating  the distance
of $v_{xyz}$ to the next voxel $v_{xyz}'$ that is in free space, i.e.\; $p(v_{xyz}' = free |
visibility) > 0$:
\vspace{-0.1cm}
\begin{equation}
p(v_{xyz} = inside | visibility) =  \frac{\min(dist(v_{xyz},v_{xyz}'), maxDist)}{maxDist},
\end{equation}
\vspace{-0.8cm}

where $dist(\cdot)$ calculates the Euclidean distance between the voxel
centers and $maxDist$, which is a predefined maximum distance, truncates this distance. Hence, this formula is closely
related to the truncated signed distance function~\cite{zach08} which is used for
example in surface extraction algorithms. 

Given the free space scores for each voxel, we can easily define the scores
for each pixel $b_{xy}$ in the 2D slice plane by averaging the scores of the
voxels:
\vspace{-0.2cm}
\begin{equation}
 p(b_{xy} = free | visibility) = 
      \sum_z \frac{p(v_{xyz} = free | visibility)}{n}, 
\end{equation}
\vspace{-0.5cm}

where $n$ is the voxel dimension in z-direction of the slice. The score that a pixel
is occupied is defined in the same way.

Finally, we obtain an optimal inside/outside labeling of the 2D slice plane by
solving an energy minimization problem where we directly use the defined
free space scores as data term. 
The pairwise terms are defined by the Potts model.

\begin{figure}
\centering
\subfigure{
\label{fig:fs:projected}
\includegraphics[width=0.2\linewidth]{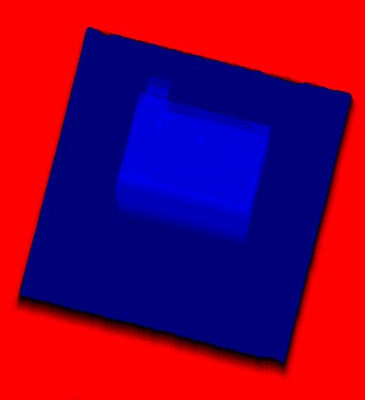}
}
\subfigure{
\label{fig:labeling}
\includegraphics[width=0.2\linewidth]{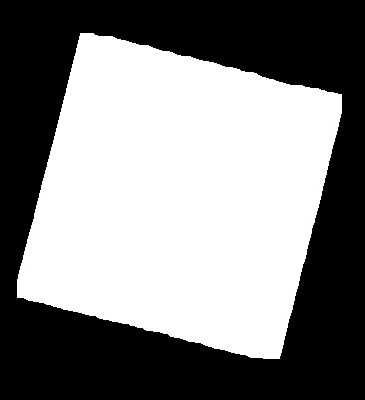}
}
\subfigure{
\label{fig:outline}
\includegraphics[width=0.2\linewidth]{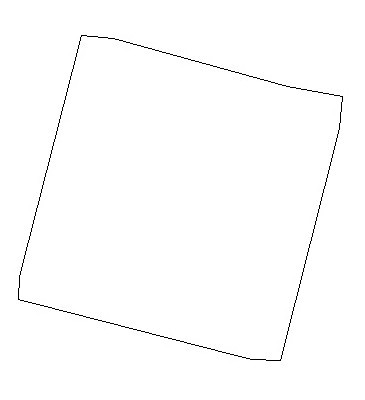}
}
\caption[Binary labeling and outline simplification of a slice.]{\textit{Slice
processing.} Left, the free space scores projected to 2D
are illustrated. All voxels within a slice are projected onto a 2D matrix and
a free space score per pixel gets calculated (blue means inside, red outside).
Using graph cut energy minimization, a
binary labeling for inside and outside is calculated (middle). On the binary labeling, outline simplification is
applied which delivers a polygonal line (right).}
\label{fig:bin_labeling}
\end{figure}

\textbf{Outline Simplification.} 
For the creation of the geometric 3D model, we just need the outline of the
inside-labeled pixels. Though, as the outline is a polygonal line including
every pixel as point and we favor simple representations, a simplification of
the polygonal outline is applied before continuing with further processing
steps. There exist several approaches for this task. We use the
\textit{Ramer-Douglas-Peucker algorithm}~\cite{douglas73} in our
implementation. The algorithm produces a polygonal line as in 
Fig.~\ref{fig:bin_labeling}, which can be easily extruded to 3D.

\vspace*{-0.4cm}
\subsection{Slice Combination}
\label{sec:3d_optimization}
\vspace{-0.2cm}

\begin{figure}
\centering
\subfigure{
\includegraphics[width=0.4\linewidth]{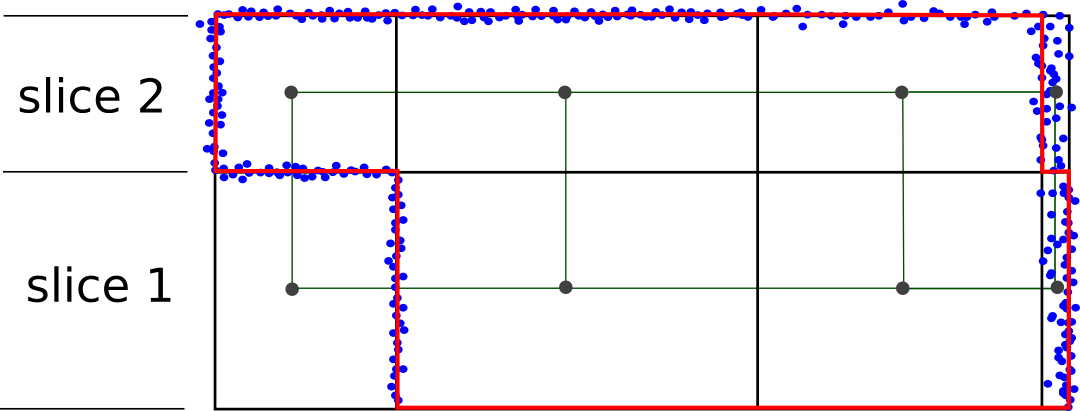}
}
\subfigure{
\includegraphics[width=0.4\linewidth]{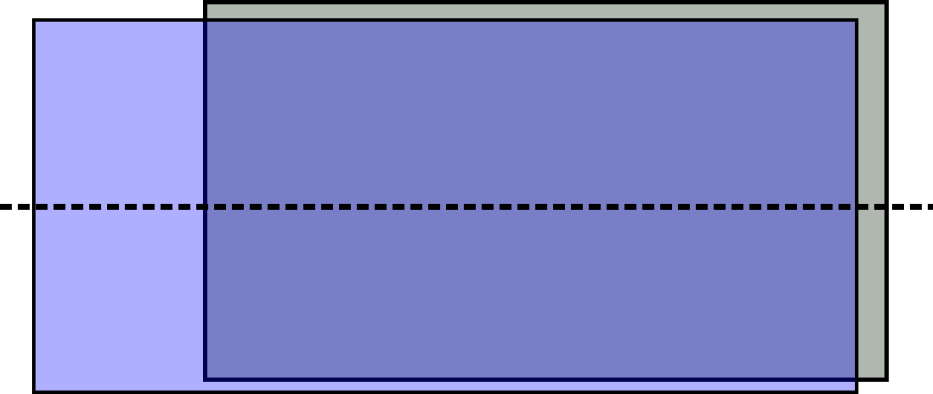}
}
\caption{\textit{Left: Volumetric Cells. }Vertical cut of a volumetric cell 
representation of a simple model consisting out of two
slices. The black lines are the volumetric cells spanned over the whole scene
and the red lines are the outlines of the extruded slices approximating the
point cloud (blue dots). A graph is spanned over the whole scene setting cells
with a shared face as neighbors (green lines). \textit{Right: Top-view of binary labeling of both slices. } As you can see, a noisy point cloud leads to slightly varying object outlines in each slice. The dashed line represents the vertical cut seen in the left image.}
\label{fig:volCells}
\end{figure}

By extruding the object outlines from each slice to 3D (as in Fig.~\ref{fig:volCells}), we can already create
a regularized 3D model. However, as you can see in the labeling masks of the two slices,
there usually exist small differences in the 2D discretization which result in 
non-smooth 3D models. Further, small
irregularities that just occur in one slice are not necessarily wanted to be
in the geometric model. Therefore, we apply
a regularization step by partitioning the whole possible occupied
space into irregularly shaped 3D volumetric cells and create again an optimal
inside/outside labeling using energy minimization.

\textbf{Volumetric Cells.}
To regularize the geometric model, we partition the model 
into irregularly shaped volumetric cells. The
concept is illustrated in Fig.~\ref{fig:volCells}. We define volumetric cells
as right prisms with triangles as base faces.

For the partitioning, we first project
all object outlines of each slice's binary segmentation to the ground plane.
Next, we apply a Constrained Delaunay Triangulation~(CDT) to this line set.
The CDT guarantees that the projected outlines remain lines in the
final triangulation. Finally, we extrude the triangles between all slice
boundaries to get an irregular space partitioning of the whole scene.

\textbf{3D Regularization.}
To get rid of unwanted irregularities between the different slices, we perform a regularization in 3D. Given the volumetric cell representation, we 
want to find an optimal inside/outside labeling for the cells. Similar to the
slice binarization, we use the visibility information to setup an energy optimization
problem that is formulated as an CRF:
\vspace{-0.2cm}
\begin{equation}
E(L) = \sum_{p \in \mathcal{I}} E_{data}(L(p)) + \lambda \sum_{p,q \in \mathcal{N}}  E_{smooth}(L(p),L(q)) ,
\end{equation}
\vspace{-0.4cm}

where $\mathcal{I}$ denotes the set of all volumetric cells, 
$\mathcal{N}$ is the neighborhood of every cell and $L$ is the (binary) labeling.
The neighborhood relation is defined by the volumetric cell complex: all 
cells that share a common face are neighbors~(see Figure~\ref{fig:volCells}).
The data terms, $E_{data}(l_p)$, are defined as
\vspace{-0.2cm}
\begin{equation}
E_{data}(l_p) = \left\{
  \begin{array}{l l}
    insideScore(p) & \quad \text{if $l_p = inside$}\\
    outsideScore(p)  & \quad \text{if $l_p = outside$}
  \end{array} \right.\ ,
\end{equation}
\vspace{-0.4cm}

where $insideScore(p)$  and $outsideScore(p)$  are the summed up free space
scores within the volumetric cell normalized by the cell height and the size
of the base face.

In our approach, the neighbor smoothness penalties from one cell to their 
neighboring cells, $E_{smooth}(p,q)$, depend on the amount of points near the 
face of adjacent cells. Using the constantly upsampled point cloud, we count the points 
which are near this face. 
With this approach, it is unlikely 
that two cells, which have a dominant structure, e.g. a wall, between them, 
get smoothed into an equally labeled group and it is likely that two cells 
without structures between them get smoothed into the same group. 
We calculate the neighbor weights as
\vspace{-0.2cm}
\begin{equation}
E_{smooth}(l_p,l_q) = \left\{
  \begin{array}{l l}
    0 & \quad \text{if $l_p = l_q$}\\
    \frac{1}{1 + \frac{\{\#\ points\ near\ face_{p,q}\}}{area\ of\ face_{p,q}}}  & \quad \text{else}
  \end{array} \right.\ ,
\end{equation}
\vspace{-0.4cm}

where $\{\#\ points\ near\ face_{p,q}\}$ is the amount of points which have a 
smaller Euclidean distance to the face than a fraction of the model size.
\\
The result has the value $ 0 < E_{smooth}(l_p,l_q) \le 1 $,
where $E_{smooth}(l_p,l_q)$ is near 0 when lots of points are near the 
adjacent face, which means there exist scene structures. In this case, 
no smoothing is wanted and due to $E_{smooth}(l_p,l_q) \approx 0$, 
the smoothness penalty is near 0. $E_{smooth}(l_p,l_q)$ is 1, when no
point is near the adjacent face, which means that the total smoothness 
penalty is completely adjusted by $\lambda$.
\\
Finally, we get the regularized 3D geometry model with a regularization level 
(a so-called level of detail) depending on $\lambda$. 

\vspace*{-0.8cm}
\section{Experiments}
\label{sec:experiments}
\vspace{-0.3cm}
The main goal of our work is to regularize meshed 3D point clouds and
simultaneously reduce the amount of data. In our experiments we show the
deviations w.r.t. geometry of the computed geometric model to the original
model and to which amount model simplification can be achieved. Our algorithm 
has been implemented in C++ using the Graph-Cut library~\cite{boykov01} to
solve energy minimization problems.

\vspace{-0.4cm}
\subsection{Evaluation Data}
\vspace{-0.2cm}
For our experiments we make use of publicly available models from the Trimble 3D 
Warehouse \cite{googleWarehouse}. As input for our workflow, we use reconstructions 
from the synthetic models. We first texture the model with a random texture 
and create virtual camera views. Then, we reconstruct the scene with SfM and
PMVS (as described in~\cite{snavely06}\cite{furukawaPMVS}) and finally mesh the point cloud using Poisson triangulation. For 
evaluation, we compare the computed geometry with the ground truth, 
which is the original synthetic model.

We also evaluate with models reconstructed from real-world image data 
using SfM, PMVS and Poisson triangulation. Although we do not have a 
correct ground truth in this case, we use the reconstruction as if it would
be the ground truth and compare the computed geometry against it.

\begin{figure}
\centering
\subfigure{
\label{fig:synthetic}
\includegraphics[width=120px]{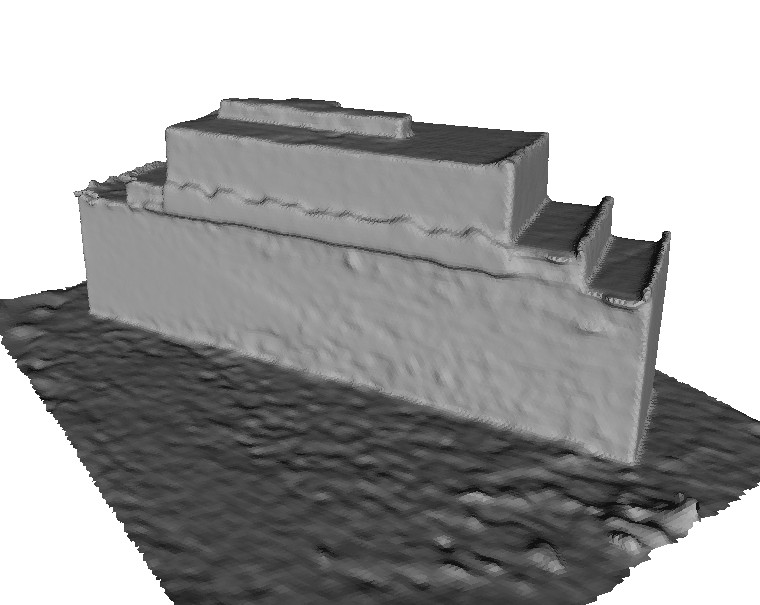}
}
\subfigure{
\label{fig:real}
\includegraphics[width=120px]{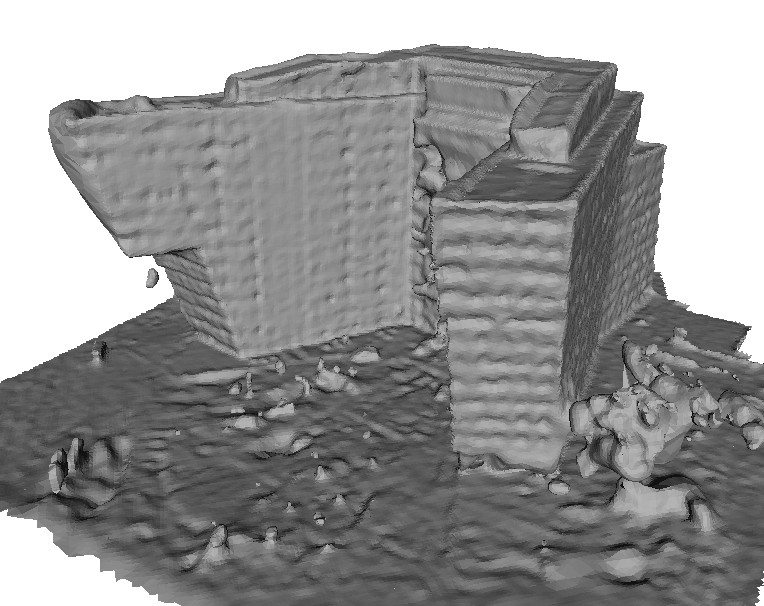}
}
\caption{\textit{Models used for evaluation}. 
Input meshes for our method obtained from image-based reconstruction.
}
\label{fig:eval_models}
\end{figure}

For both types, we use models from buildings which meet our 
requirements (i.e., horizontal planar structures and orthogonal vertical surfaces) 
up to a certain degree~(Fig.~\ref{fig:eval_models}). 

\vspace{-0.4cm}
\subsection{Evaluation Metrics}
\vspace{-0.2cm}
As we want to know how well the computed geometry approximates the model and which degree of simplification could be achieved, we define an error measure and a regularization measure.

\textbf{Error Measure.}
In our evaluation, we make use of an error measure which approximates the perceived 
error of humans by comparing the computed geometry and the ground truth model.
For this, we use the Dice score~\cite{dice45} of the two models backprojected 
into all camera views. The Dice score relates the area of the two projected
segments with the area of their mutual overlap. It is 1 if the two segments
are completely identical and 0 if there is no overlap. To get a score for the
whole model, we use the mean Dice score of all cameras. Using this measure, 
the 3D problem gets translated into a 2D problem 
consisting out of segmentation masks, which can be evaluated much easier.

\textbf{Regularization Measure.}
As a main goal is to regularize the input data, we have to define a measure 
for the degree of regularization. We project the outlines of each slice 
(i.e., the vertical faces of the geometric model) onto the ground plane and
count the number of unique lines as a complexity score. The more lines exist, 
the less regularized is the model.

\vspace{-0.5cm}
\subsection{Results}
\begin{figure}
\centering
\subfigure[Ground Truth]{
\includegraphics[width=0.3\linewidth]{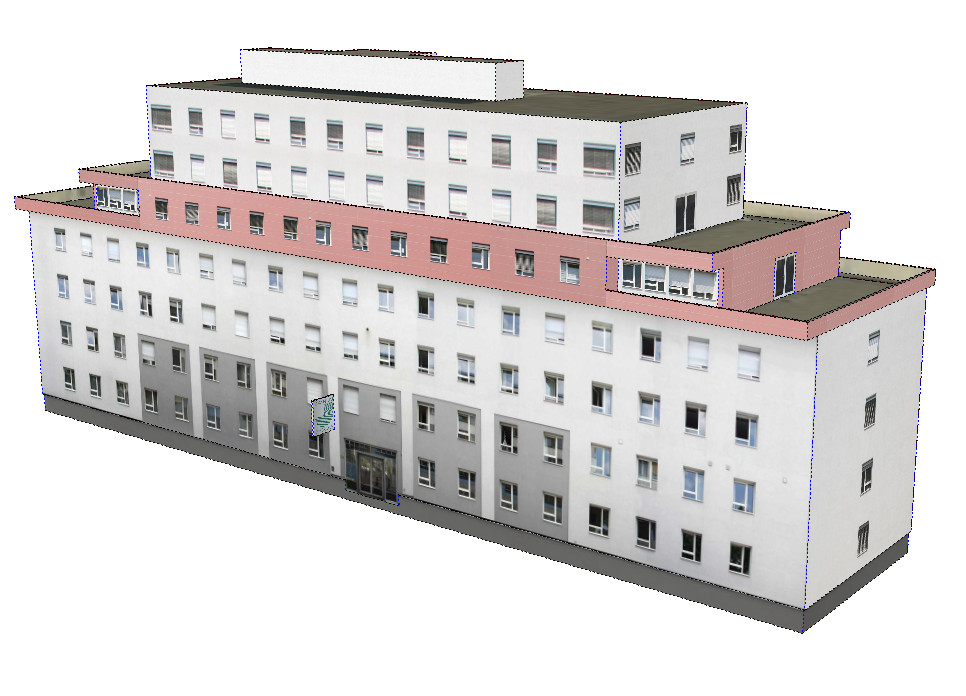}
\label{fig:synthetic_res:gt}
}
\subfigure[$d = 25$, $\lambda = 4.5,\ \ \ $\hspace{\textwidth} $score = 0.9702$, $compl. = 50$]{
\includegraphics[width=0.3\linewidth]{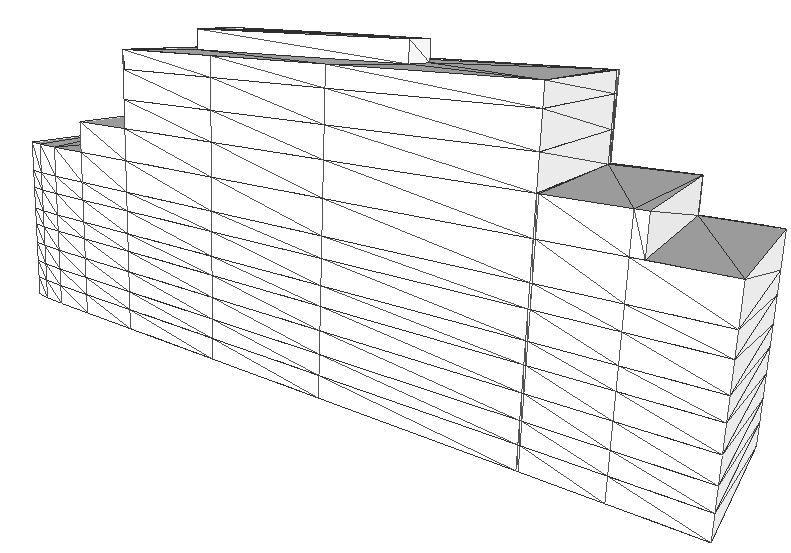}
\label{fig:synthetic_res:d}
}
\subfigure[$d = 10, \lambda = 1.0,\ \ \ $\hspace{\textwidth} $score = 0.9709$, $compl. = 41$]{
\includegraphics[width=0.3\linewidth]{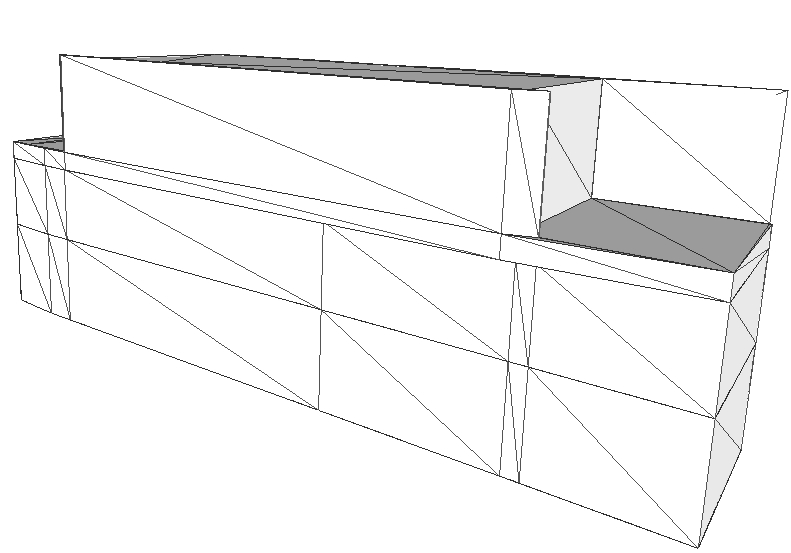}
\label{fig:synthetic_res:a}
}
\subfigure[$d = 25, \lambda = 2.5,\ \ \ \ \ \ \ \ $\hspace{\textwidth} $score = 0.9703$, $compl. = 212$]{
\includegraphics[width=0.3\linewidth]{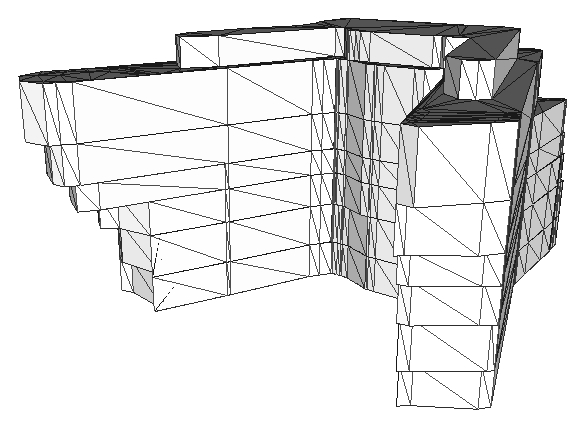}
}
\subfigure[$d = 35, \lambda = 2.5,\ \ \ \ \ \ \ \  $\hspace{\textwidth} $score = 0.9770$, $compl. = 472$]{
\includegraphics[width=0.3\linewidth]{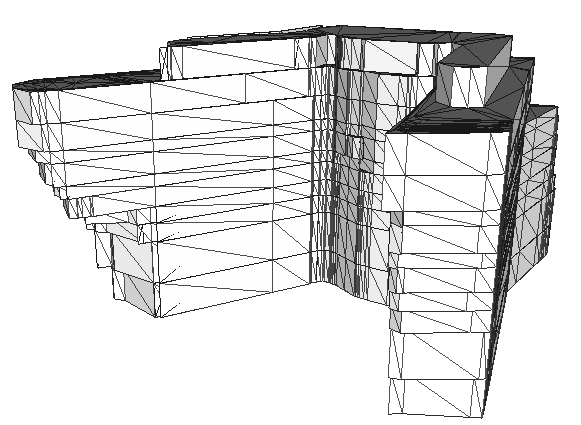}
}
\caption{\textit{Results.} In the first row, you can see geometric models created from 
the synthetic model and in the second row, models created from the real-world
model with varying parameter settings.
}
\label{fig:synthetic_res}
\end{figure}

\vspace{-0.4cm}
In the first row of Fig.~\ref{fig:synthetic_res} one can see results of the synthetic model 
with different parameter settings, the corresponding mean Dice score and the
complexity measure. 
The Dice score and the complexity of the model are directly related to the
parameter $\lambda$ and to the mean shift bandwidth. 
As a lesser amount of slices directly results in a big decrease of the
regularization measure, models calculated with a higher bandwidth implicitly
have a lower complexity score.
Though, a
lower mean shift bandwidth with much more smoothing can lead to the same 
regularization level as a higher bandwidth with lesser smoothing. More
precisely, a too low bandwidth is not as critical as a too high bandwidth, as
it can be compensated by a higher smoothing. 

The second model used for our evaluation is a 3D model reconstructed from 
real-world image data and therefore includes more noise and irregularities.
Further, the scene geometry is not perfectly suited for our approach as some 
vertical surfaces are not orthogonal to the ground plane. However, as you 
can see in the second row of Fig.~\ref{fig:synthetic_res}, a small mean shift bandwidth leads also 
to an acceptable approximation of sloped structures.

In the results of both models, one can see that the incorporated noise does
not have a big influence on the model quality. 

For each of the resulting geometric models, we needed approximately 16 to 17 
minutes of computation time on a Intel Xenon X5675 with 16 GBs RAM. However, 
as most of the computation time is needed for computing the free space score
voxel grid, the time can be drastically reduced by using a smaller 
voxel grid. Further, the recomputation of results for varying values of $\lambda$ can be done within seconds. 

\vspace{-0.5cm}
\section{Conclusions and Future Work}
\label{sec:conclusions}
\vspace{-0.3cm}
In this paper we proposed a novel approach for extracting geometric structures
out of meshed point clouds reconstructed from image-based reconstruction
methods. While similar techniques are used on laser scan data, we have
shown that our approach can also handle noise and small irregularities due to
errors in the reconstruction process.

Future work includes the detection of an optimal value for $d$ depending on the scene structure
and the integration of additional scene information retrieved from the 2D
images.
\vspace{-0.4cm}

\section*{Acknowledgments}
\vspace{-0.3cm}
This work was supported by the Austrian Research Promotion Agency (FFG) within 
the CONSTRUCT research project funded by Siemens AG Austria. We want to thank
the Siemens team in Graz for discussion and for providing 
data used for our experiments.

\vspace*{-0.4cm}
\bibliography{myrefs}
\end{document}